\documentclass[graybox]{svmult}

\usepackage{type1cm}        % activate if the above 3 fonts are
\usepackage{makeidx}         % allows index generation
\usepackage{graphicx}        % standard LaTeX graphics tool
\usepackage{multicol}        % used for the two-column index
\usepackage[bottom]{footmisc}% places footnotes at page bottom

\usepackage{newtxtext}       % 
\usepackage[varvw]{newtxmath}       % selects Times Roman as basic font

\makeindex             % used for the subject index
\begin{document}

\title*{Stacked Tensegrity Mechanism for Medical Application}
\author{Dhruva Khanzode, Ranjan Jha, Emilie Duchalais and Damien Chablat}
\institute{Dhruva Khanzode, Ranjan Jha \at BioMedical Application Division,  CSIR- Central Scientific Instruments Organisation, Chandigarh, India, Academy of Scientific and Innovative Research (AcSIR), Ghaziabad, Uttar Pradesh 201002, India
\email{khanzode.dhruva@csio.res.in, ranjan.jha@csio.res.in}
\and Damien Chablat \at Laboratoire des Sciences du Num\'erique de Nantes, UMR CNRS 6004, Nantes, France \email{damien.chablat@cnrs.fr}
\and Emilie Duchalais \at CHU Nantes - Centre hospitalier universitaire de Nantes, France
\email{emilie.dassonneville@chu-nantes.fr}}

\maketitle

\abstract{In this article a multi-segmented planar tensegrity mechanism was presented. This mechanism has a three-segment structure with each segment residing on top of another. The size of the segments may decrease proportionally from base to top, resulting in a tapered shape from base to tip like an elephant trunk.
The system was mechanically formulated as having linear springs and cables functioning as actuators. The singularities, as well as the stability of the parallel mechanism, were analyzed  by using the principle of minimum energy. Optimization was also done to obtain the greatest angular deflection for a segment according to a ratio between the size of the base and the moving platform of the robotic system. The result of this work is a family of mechanisms that can generate the same workspace for different stability properties.}
%%%%%%%%%%%%%%%%%%%%%%%%%%%%
\section{Introduction}
\label{sec:1}
%%%%%%%%%%%%%%%%%%%%%%%%%%%%
Robotics is one of the most rapidly emerging branches in the world of technology. From heavy-duty applications to the most sensitive microscopic application, robotic systems are present everywhere. Therefore, with the rise of applications of robotic systems, grew the demand for facilitating critical tasks and requirements of high dexterity applications. To fulfill these demands, a separate branch of robotics emerged coined as flexible robotic. A field in which manipulators and actuators of hyper redundant mechanisms were developed. Hyper redundant systems are robotic systems having a very high number of degree of freedom \cite{yesh_1}. These flexible robotic systems are very versatile in their applications. These flexible robotic systems are generally employed at a place with convoluted workspace and restricted maneuverability \cite{degani}. Such kind of flexible robotic systems can be abundantly seen as probing and inspection drones, surgical robots, military UGVs, etc \cite{wang_1,hwang_1}.
The flexible robotic systems also called snake/worm-like robotic systems can be divided into two groups, based on their mechanical structure, discrete systems, and continuous systems. Discrete systems have finite discrete rigid elements joint together \cite{yesh_2} while the continuous systems have soft or compliant materials used in the robotic system \cite{suzumori}. But in recent times, the line between these two categories has been blurred as newer designs being developed have both of the features. Another classification of hyper redundant robotic systems can be made based on how the robot is used. It can be classified either as a manipulator or a locomotor. If a robotic system is attached to a fixed base, then it can be called a manipulator \cite{xu_1}, but if the system can  move freely, (such as a work or a snake), then it is to be called a locomotor \cite{loco}.\\
The flexible robotic systems are generally tendon actuated, pneumatically, or hydraulically actuated or are based on the shape memory feature of the material. Tendon-based mechanisms are one of the most used mechanisms for flexible manipulators. The manipulators are actuated by pushing or pulling a tendon to facilitate bending or curvature \cite{li_1}. These tendons-driven mechanisms are sometimes also referred to as tensegrity mechanisms because of the design principle wherein compression and expansion of lengths of opposite tendons takes place simultaneously \cite{furet r1, wenger r2}. A study was conducted to analyze and compare the kinematic properties of discrete tendon-driven flexible robots, continuous tendon-driven flexible robots, and concentric tube manipulators \cite{li_2}. The results of this study indicated that in the case of dexterity, the discrete system was the finest followed by continuous and concentric tube type manipulators. From a scaling point of view, concentric tube manipulators are better for miniaturization whereas discrete tendon-driven systems were more suitable for scaling up \cite{le_1}.\\
%% dhruva edit follows:
This flexible robotic system has found some very interesting applications in the field of medical sciences \cite{shang}. Even before the conceptualization of robotic surgeries, endoscopes were used to investigate the insides of the human gastrointestinal tract, and it was done manually \cite{haber,abbott}. With the medical science turning into minimally invasive surgeries (MIS), several surgical tools were required to be redesigned for MIS such as clippers, scalpels, cauterizers, catheters, etc \cite{roh,hu_1}. One such essential surgical tool is the surgical stapler \cite{Akopov2021}, which is generally used to avoid any spillage of intestine content onto the wounds of the patient and prevent any infection during the surgery. Commonly five kinds of staplers are used, namely TA, Thoracic-Abdominal; GIA, Gastro-Intestinal Anastomosis; Endo GIA, Endoscopic Gastro-Intestinal Anastomosis; EEA, End-End Anastomosis and Skin Stapler. The staplers are generally equipped with a separating knife to cut the tissues after the stapling is done. Endo GIA staplers are most commonly used for MIS. The Endo GIA staplers are available in three forms: passive articulated wrist type (PAW) \cite{Bolanos1997}, active articulated wrist type (AAW) \cite{Milliman1999}, and radial reload type staplers (RR). The RR has a fixed ``U'' shaped jaw which requires a huge incision into the abdomen for insertion. Therefore there is a dire need for a surgical stapler that can enter through the laparoscopic openings but can work as an RR-type stapler inside the body. Our study aims to design a stapler for laparoscopic rectal cancer surgery where conventional tools cannot be easily accessed. Two mechanisms will have to be positioned on each side of the colon to insert two lines of staples and then carry out its cutting.

%Even before the conceptualisation of robotic surgeries, endoscopes were used to investigate the insides of human gastro-intestinal tract, and it was done manually \cite{haber}. With the development of surgical robots, the robotic endoscopes were developed to assist the surgeon performing the robotic surgery visually. Since then, this technology has advanced exponentially \cite{abbott} to an extent such that nowadays minimally invasive single port laparoscopic surgeries are possible due to the replacement of 2 DOF tendon-driven revolute jointed wrist with hyper redundant flexible manipulators \cite{roh}. Several distal tip active catheters are also present for cardiac ablation procedures \cite{hu_1}. The aim of our study is to design a stapler for laparoscopic rectal cancer surgery where conventional tools cannot be easily accessed. Two mechanisms will have to be positioned on each side of the colon to insert two lines of staples and then carry out its cutting.

The article is organized as follows. In section 2, the kinematics of the studied mechanism is presented. Section 3 reports the different computational results to study the singularities associated with the mechanism, then in section 4, the study of stability based on the principle of minimum energy, and finally, in section 5, an optimization of the dimensions of the mechanism is presented.
%%%%%%%%%%%%%%%%%%%%%%%%%%%%%
\section{Mechanism Design}
\begin{figure}
        \centering
        \includegraphics[width=12cm]{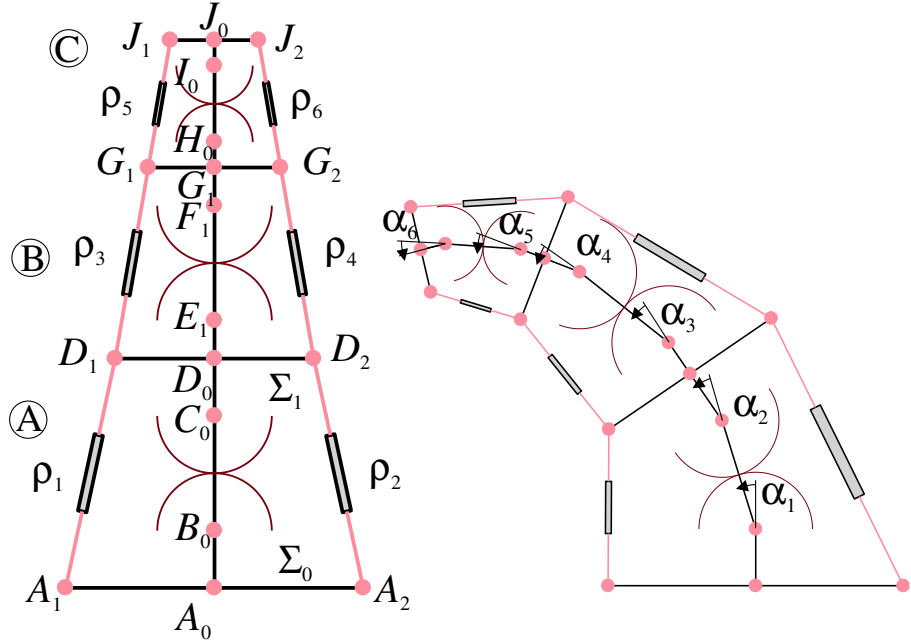}
        \caption{The tensegrity  mechanism understudy with three segments stacked, named A, B, C for $\alpha_i=0$ (home pose) in the left and $\alpha_i=0.3$ in the right}
        \label{fig:robot_general}
\end{figure}

%%%%%%%%%%%%%%%%%%%%%%%%%%%%%
\label{sec:2}
In this article, the robotic system designed is a multi-segment planar robotic system. As depicted in Fig.~\ref{fig:robot_general}, the robotic system comprises of 3 trapezoidal segments A, B, and C stacked one above the other. Each segment may have a larger base plate and a smaller moving plate. Both the plates are connected  by a central serial spine linkage, comprising of 3 links. The system is only composed of 3 sections to facilitate the position control of the end effector. Since the system is being controlled by only 2 tendons, hence it is under-actuated. Therefore, not more than 3 segments can be controlled effectively. 

For example, in section A, the central spinal linkages are $A_{0}B_{0}$, $B_{0}C_{0}$ and $C_{0}D_{0}$. The link $A_{0}B_{0}$ and $C_{0}D_{0}$ are rigidly fixed and are perpendicular to their respective plate $A_{1}A_{2}$ and $D_{1}D_{2}$. These central spinal links are connected by two revolute joints with identical rotation angles. To achieve this coupling, the joints can have a sliding surface, similar to a knee joint in humans, or can also have an X-shape tensegrity module \cite{furet r1, wenger r2}, or a gear train to couple the movement of the two revolute joints.

The base plate and the moving plates are also attached with the help of two cables $\rho_1$ and $\rho_2$, present on either side of the central spine and two springs of stiffness $k_1$ and $k_2$ between ($A_1$ $B_1$) and ($A_2$ $B_2$), respectively.

The contraction of the cables will stimulate angular displacement in the revolute joints of the central spine and replication of this phenomenon in each of the three segments will eventually facilitate the bending of the robotic structure. The length of the cables is measured as $\rho_i$ and the angular deviation of the revolute joints is measured as $\alpha_i$ where
\begin{equation}
    \alpha_{1} = \alpha_{2},
    \quad \alpha_{3} = \alpha_{4}, 
    \quad\alpha_{5} = \alpha_{6},
\end{equation}
because of the coupling inside each segment.

The fixed coordinate frame of the base is represented by $\Sigma_0$ with the origin at $A_0$. The distances between the coordinates of each points  are $||{\bf a}_1 - \bf{a}_0||=l_1$, $||{\bf a}_2 - \bf{a}_0||=l_1$, $||{\bf b}_1 - \bf{a}_0||=h_1$, $||{\bf c}_0 - \bf{b}_0||=h_2$, $||{\bf d}_0 - \bf{c}_0||=h_3$, $||{\bf d}_1 - \bf{d}_0||=l_2$ and $||{\bf d}_2 - \bf{d}_0||=l_2$.
The vector co-ordinates for the base mounting points are given by
\begin{equation}
    {\bf a}_1=[-l_1,0] \quad {\bf a}_2=[l_1,0]
\end{equation}
The kinematic chain between the base and the platform of the first segment is defined by the following points
\begin{equation}
    {\bf b}_0= \left[\begin{array}{c}
                  0 \\
                 h_{1} 
                \end{array}\right],~
    {\bf c}_0= \left[\begin{array}{c}
               -h_{2} \sin \! \left(\alpha_{1}\right) \\
               h_{1}+h_{2} \cos \! \left(\alpha_{1}\right) 
               \end{array}\right],~
    {\bf d}_0= \left[\begin{array}{c}
               -h_{2} \sin \! \left(\alpha_{1}\right)-h_{3} \sin \! \left(2 \alpha_{1}\right) \\
               h_{1}+h_{2} \cos \! \left(\alpha_{1}\right)+h_{3} \cos \! \left(2 \alpha_{1}\right) 
               \end{array}\right]
\end{equation}
The moving co-ordinate frame of the first segment is represented by $\Sigma_1$ with its origin at $D_0$. The spring mounting points are represented by $D_1$ and $D_2$.
\begin{eqnarray}
{\bf d}_1&=&\left[\begin{array}{c}
\left(-2 h_{3} \cos \! \left(\alpha_{1}\right)-h_{2}\right) \sin \! \left(\alpha_{1}\right)-2 l_{2} \left(\cos^{2}\left(\alpha_{1}\right)\right)+l_{2} 
\\
 2 h_{3} \left(\cos^{2}\left(\alpha_{1}\right)\right)+\left(-2 l_{2} \sin \! \left(\alpha_{1}\right)+h_{2}\right) \cos \! \left(\alpha_{1}\right)+h_{1}-h_{3} 
\end{array}\right] \\
{\bf d}_2&=&\left[\begin{array}{c}
\left(-2 h_{3} \cos \! \left(\alpha_{1}\right)-h_{2}\right) \sin \! \left(\alpha_{1}\right)+2 l_{2} \left(\cos^{2}\left(\alpha_{1}\right)\right)-l_{2} 
\\
 2 h_{3} \left(\cos^{2}\left(\alpha_{1}\right)\right)+\left(2 l_{2} \sin \! \left(\alpha_{1}\right)+h_{2}\right) \cos \! \left(\alpha_{1}\right)+h_{1}-h_{3} 
\end{array}\right]
\end{eqnarray}
The inverse kinematic model for a segment is used to determine the length of the springs and the cables between the base and the moving platform and the moving platform. The equations are given by 
\begin{equation}
    ||{\bf a}_{1}-{\bf d}_{1}||=\rho_{1}, \quad   
    ||{\bf a}_{2}-{\bf d}_{2}||=\rho_{2}
\end{equation}
These two equations can also be written as, 
\begin{eqnarray}
\left(\left(-2 h_{3} \cos \! \left(\alpha_{1}\right)-h_{2}\right) \sin \! \left(\alpha_{1}\right)-2 l_{2} \left(\cos^{2}\left(\alpha_{1}\right)\right)+l_{2}+l_{1}\right)^{2} \nonumber\\
+\left(2 h_{3} \left(\cos^{2}\left(\alpha_{1}\right)\right)+\left(-2 l_{2} \sin \! \left(\alpha_{1}\right)+h_{2}\right) \cos \! \left(\alpha_{1}\right)+h_{1}-h_{3}\right)^{2}= \rho_1^2 \label{closed_loop_1} \\
\left(\left(-2 h_{3} \cos \! \left(\alpha_{1}\right)-h_{2}\right) \sin \! \left(\alpha_{1}\right)+2 l_{2} \left(\cos^{2}\left(\alpha_{1}\right)\right)-l_{2}-l_{1}\right)^{2} \nonumber\\
+\left(2 h_{3} \left(\cos^{2}\left(\alpha_{1}\right)\right)+\left(2 l_{2} \sin \! \left(\alpha_{1}\right)+h_{2}\right) \cos \! \left(\alpha_{1}\right)+h_{1}-h_{3}\right)^{2}= \rho_2^2
\end{eqnarray}
%%%%%%%%%%%%%%%%%%%%%%%%%%%%%%%%%%%%%%
\section{Singularity Analysis}
%%%%%%%%%%%%%%%%%%%%%%%%%%%%%%%%%%%%%%
For determining the singularities of the multi-segment planar mechanism, only  section A was taken into consideration. As all other sections are similarly constructed, analysis of singularities in one section will lead to similar results in all other sections, and sections B and C will also have similar singularities.

Two closed-loop mechanisms can be described by ($A_1$, $A_0$, $B_0$, $C_0$, $D_0$, $D_1$) and ($A_2$, $A_0$, $B_0$, $C_0$, $D_0$, $D_2$). For the first closed-loop mechanism, the singular configurations with $\alpha_1$ as Cartesian values and  $\rho_1$ as the input values are computed by differentiating with respect to time the Eq.~\ref{closed_loop_1}, as follow,

\begin{eqnarray}
&& -8 h_{3}^{2} \cos^{3}\left(\alpha_{1}\right) \sin \! \left(\alpha_{1}\right)+8 h_{3}^{2} \cos \! \left(\alpha_{1}\right) \sin \! \left(\alpha_{1}\right)-8 l_{2}^{2} \cos^{3}\left(\alpha_{1}\right) \sin \! \left(\alpha_{1}\right)\nonumber\\
&& +8 l_{2}^{2} \cos \!\left(\alpha_{1}\right) \sin \! \left(\alpha_{1}\right)-4 h_{3} \sin \! \left(\alpha_{1}\right) \cos^{2}\left(\alpha_{1}\right) h_{2}-4 h_{3} \sin^{3}\left(\alpha_{1}\right) h_{2}\nonumber\\
&& -4 h_{3} \cos^{2}\left(\alpha_{1}\right) l_{1}+4 h_{3} \sin^{2}\left(\alpha_{1}\right) l_{1}-4 l_{2} \cos^{2}\left(\alpha_{1}\right) h_{1}+4 l_{2} \sin^{2}\left(\alpha_{1}\right) h_{1}\nonumber\\
&& -8 h_{3}^{2} \sin^{3}\left(\alpha_{1}\right) \cos \! \left(\alpha_{1}\right)-2 h_{2} \cos \! \left(\alpha_{1}\right) l_{2}-2 h_{2} \cos \! \left(\alpha_{1}\right) l_{1}+8 l_{2} \cos \! \left(\alpha_{1}\right) l_{1} \sin \! \left(\alpha_{1}\right)\nonumber\\
&& -8 l_{2}^{2}\sin^{3}\left(\alpha_{1}\right) \cos \! \left(\alpha_{1}\right)-8 h_{3} \cos \! \left(\alpha_{1}\right) h_{1} \sin \! \left(\alpha_{1}\right)-2 h_{2} \sin \! \left(\alpha_{1}\right) h_{1}\nonumber\\
&& +2 h_{2} \sin \! \left(\alpha_{1}\right) h_{3}=0
\end{eqnarray}

The singularities are the roots of a 4th-degree equation. When $l_1 \neq l_2$, only a numerical method allows us to calculate them. In our case, we use the ``RootFinding:-Isolate'' function of Maple which computes all the roots after a substitution by the half-angle of $\alpha_1$ to obtain an algebraic equation \cite{rouillier}.

For the second closed-loop, the singularity locus is for opposite values of $\alpha_1$. For each closed loop, there can be up to four singular positions. The absolute value of the smallest angle $\alpha_i$ is called $\alpha_{sing}$ and it represents the largest travel that the mechanism can achieve. This value must be maximized.

\begin{figure}
        \centering
        \includegraphics[width=10cm]{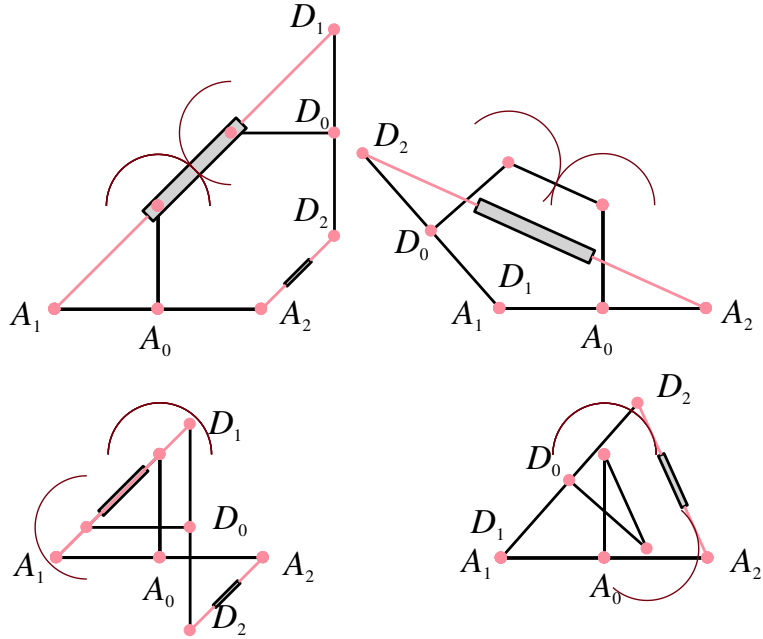}
        \caption{Four singular configurations for $h_{1} = 1, h_{2} = 1, h_{3} = 1, l_{1} = 1$ and $l_{2} = 1$}
        \label{fig:robot_sing}
\end{figure}

As an example, for $h_1=1$, $h_2=1$, $h_3=1$, $l_1=1$, $l_2=1$, the singularity locus are 
\begin{equation}
\alpha_1=-\frac{\pi}{4}, \alpha_1=\frac{3 \pi}{4}, 
\alpha_1=\arctan \! \left(\frac{\frac{1}{4}+\frac{\sqrt{7}}{4}}{-\frac{1}{4}+\frac{\sqrt{7}}{4}}\right), \alpha_1=\arctan \! \left(\frac{\frac{1}{4}-\frac{\sqrt{7}}{4}}{-\frac{1}{4}-\frac{\sqrt{7}}{4}}\right)-\pi
\end{equation}
Only two singular configurations are close to the home pose as shown in Fig.~\ref{fig:robot_sing}. The smallest absolute value of $\alpha_1$ defines the range of motion of the segment in both directions.
%%%%%%%%%%%%%%%%%%%%%%%%%%%%%%%%%%
\section{Stability  Analysis}
\label{sec:3}
%%%%%%%%%%%%%%%%%%%%%%%%%%%%%%%%%%
The stability of the robotic system is one of the major components in its performance evaluation \cite{behzadipour st1}. Generally, when a system is in equilibrium and is displaced by an external force if the system comes back to its equilibrium state, it is called a stable equilibrium. If the system does not return to equilibrium, it is considered an unstable system and if the system does not affect itself, then the system is known as a neutral system \cite{chadefaux st2}.
A robotic system is always desired to be a stable system in its operational capacity. If a robotic system becomes unstable at any point in its trajectory, then it may cause substantial errors during its control \cite{wanda}.
For this multi-segment planar robotic system, the stability is evaluated by using the principle of minimum internal energy. 

For section A, two springs are connected between $A_1D_1$ and $A_2D_2$ with a no-load length equal to $l_0$. This length is chosen as 40\% of the length in the home pose. This value will be proportional to the size of the second and third segments if we want to characterize  the stability of the complete mechanism. 
 
For segment A,  the energy $E$ for a given $\alpha$ inside the mechanism is
\begin{equation}
    E= \frac{1}{2}\left(k_1(\rho_1-l_0)^2+k_2(\rho_2-l_0)^2 \right)
\end{equation}
To compare two mechanisms, we introduce the total energy as $E_t=\int_{-\alpha_{sing}}^{\alpha_{sing}} E$.

\begin{figure}
        \centering
        \includegraphics[width=12cm]{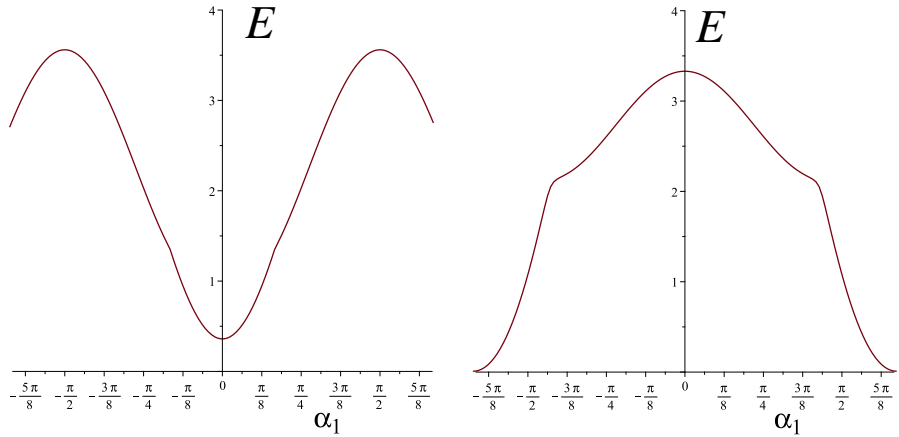}
        \caption{Variation of the energy as a function of $\alpha_1$ for a stable home pose for $h_{1} = 0$, $h_{2} = 1$, $h_{3} = 0$, $l_{1} = 1$, $l_{2} = 1$, $k = 1$ (left) and unstable home pose for $h_{1} = 1$, $h_{2} = 1$, $h_{3} = 1$, $l_{1} = 1$, $l_{2} = \frac{1}{2}$, $k = 1$}
        \label{fig:robot_energy}
\end{figure}

Depending on the sizes of the robot, several stability schemes can appear either stable around $\alpha=0$ up to the singularities (Fig.~\ref{fig:robot_energy}(left)) or unstable in the position for $\alpha=0$ (Fig.~\ref{fig:robot_energy}(right)).

%%%%%%%%%%%%%%%%%%%%%%%%%%%
\section{Optimization of One Segment}
%%%%%%%%%%%%%%%%%%%%%%%%%%%
The objective of our optimization is to find the design parameters that allow the greatest angular deflection and then to study the stability of the solutions found. A ratio $\lambda$ between the base and the platform is also studied to see if elephant trunk structures, with sections of decreasing sizes, are relevant \cite{yang_1}.

So, the objective function is defined as
\begin{equation}
    f({\bf x})= \alpha_{sing}  \rightarrow \max
\end{equation}
where ${\bf x}$ is the set of the design parameters $[h_1, h_2, h_3, l_1, \lambda]$. To simplify the optimization, we set $h_3=h_1$, $l_2= \lambda l_1$ and we defined the following constraints:
\begin{eqnarray}
0<l_1<4.5 \quad 0 \leq h_1 \leq 1 \quad 0 \leq h_2 \leq 2 \quad 1/20 \leq \lambda \leq 1
\end{eqnarray}
The solution to this optimization problem can be achieved by several methods. As the dimension of the problem is small, a discretization of the parameter space has been performed to find the optimal solutions.

\begin{figure}
        \centering
        \includegraphics[width=5.5cm]{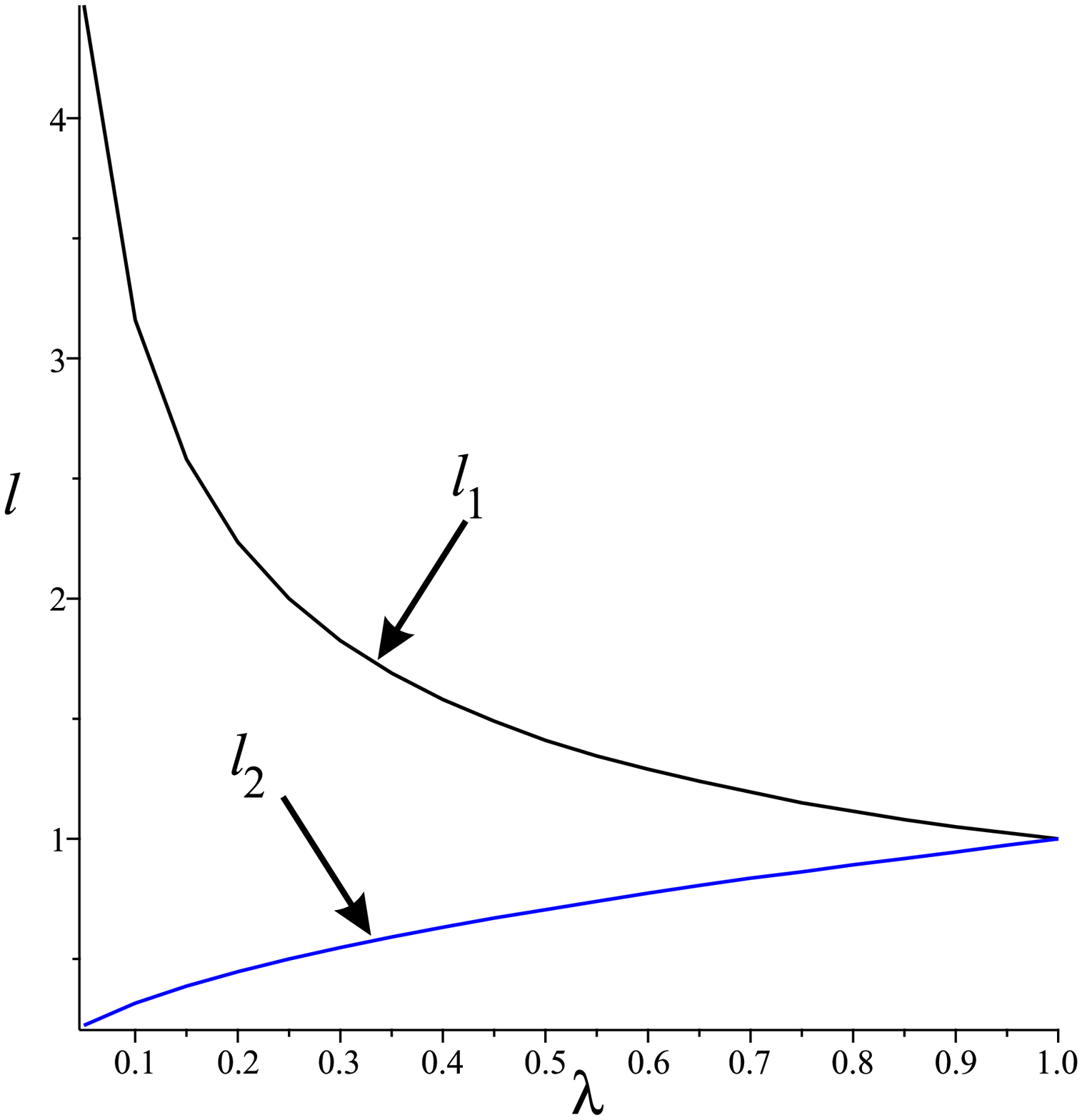}
         \includegraphics[width=6cm]{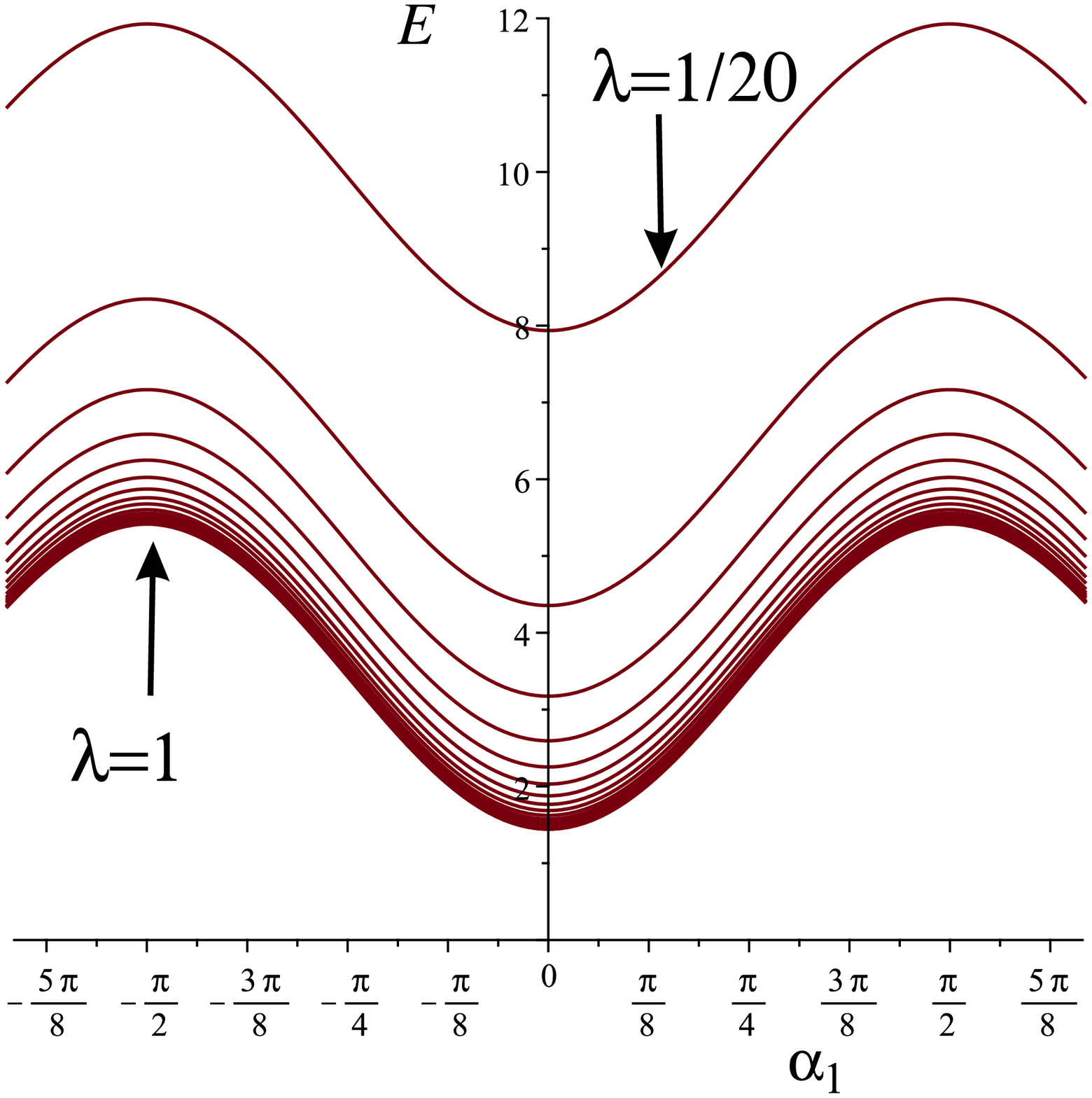}
        \caption{Variation of $l_1$ in black and $l_2$ in blue as a function of $\lambda$ (left) and the stability analysis of the optimal solutions as a function of $\lambda$}
        \label{fig:robot_l}
\end{figure}

The results of the optimization show that the maximum value of $\alpha$ is $\pi/2$ and this for any value of $\lambda$ and for $l_2=2$, i.e. the maximum bound, and for $h_1=0$ for the minimum bound. Figure~\ref{fig:robot_l} shows the evolution of $l_1$ and $l_2$ as a function of $\lambda$. 

For all solutions, the energy has its maximum value for $\alpha_{sing}$ and has a local minimum for $\alpha=0$.
If we observe the value of $E(\alpha=0)$ and $E(\alpha=\pi/2)$, the difference is always the same. The total energy decreases when $\lambda$ tends to 1. This means that the actuation forces will be smaller when $\lambda=1$. Conversely, the smaller $\lambda$ is, the greater the external forces must be to move the mechanism from its home position. Moreover, to have a regular stack of modules, module C is smaller than module B which is smaller than module A if $\lambda$ is different from 1.
%%%%%%%%%%%%%%%%%%%%%%%%%%%%%%%%%%%
\section{Conclusions and Future Work}
\label{sec:4}
%%%%%%%%%%%%%%%%%%%%%%%%%%%%%%%%%%%
To summarise, in this article a multi-segmented planar tensegrity mechanism was designed consisting of a stack of three tensegrity mechanisms. The mechanism was mechanically formulated as having linear springs and cables for actuation. The singularities of the parallel mechanism were analyzed as well as the stability by using the principle of minimum energy. Optimization was done to obtain the greatest angular deflection according to a ratio between the size of the base and the platform.  

We observe that the angle $\alpha$ can vary between $-\pi/2$ and $\pi/2$, that all optimal solutions are stable up to the singularity, and that when $\lambda=1$, the total energy in a module is minimal. 

Further research is in progress to control the three modules with only two cables going from the base to the third segment and to know the stability of the mechanism against external forces. We also need to estimate the radii of curvature required to encompass the patients' colon from CT scan studies and encapsulate the mechanism with an elastomer with the clamps and knife to section the colon.
%\begin{acknowledgement}
%If you want to include acknowledgments of assistance and the like at the end of an %individual chapter please use the \verb|acknowledgement| environment -- it will %automatically be rendered in line with the preferred layout.
%\end{acknowledgement}
%


\begin{thebibliography}{99.}%

\bibitem{yesh_1}A. Yeshmukhametov, K. Koganezawa, and Y. Yamamoto, “A novel discrete wire-driven continuum robot arm with passive sliding disc: Design, kinematics and passive tension control,” \textit{Robotics}, vol. 8, no. 3, pp. 1--18, 2019, doi: 10.3390/ROBOTICS8030051.

\bibitem{degani}A. Degani, H. Choset, B. Zubiate, T. Ota, and M. Zenati, “Highly articulated robotic probe for minimally invasive surgery,”\textit{ Proc. 30th Annu. Int. Conf. IEEE Eng. Med. Biol. Soc. EMBS’08 - "Personalized Healthc. through Technol.}, no. May, pp. 3273--3276, 2008, doi: 10.1109/iembs.2008.4649903.

\bibitem{wang_1}H. C. Wang, S. H. Cui, Y. Wang, and C. L. Song, “A hybrid electromagnetic and tendon-driven actuator for minimally invasive surgery,” \textit{Actuators}, vol. 9, no. 3, 2020, doi: 10.3390/act9030092.

\bibitem{hwang_1}M. Hwang and D. S. Kwon, “K-FLEX: A flexible robotic platform for scar-free endoscopic surgery,” \textit{Int. J. Med. Robot. Comput. Assist. Surg.}, vol. 16, no. 2, 2020, doi: 10.1002/rcs.2078.

\bibitem{yesh_2}A. Yeshmukhametov, K. Koganezawa, and Y. Yamamoto, “Design and Kinematics of Cable-Driven Continuum Robot Arm with Universal Joint Backbone,” 2018 \textit{IEEE Int. Conf. Robot. Biomimetics, ROBIO 2018}, pp. 2444--2449, 2018, doi: 10.1109/ROBIO.2018.8665186.

\bibitem{suzumori}K. Suzumori, S. Iikura, and H. Tanaka, “Development of flexible microactuator and its applications to robotic mechanisms,” \textit{Proc. - IEEE Int. Conf. Robot. Autom.}, vol. 2, no. April, pp. 1622--1627, 1991, doi: 10.1109/robot.1991.131850.

\bibitem{xu_1}K. Xu, J. Zhao, S. Member, and M. Fu, “Development of the SJTU Unfoldable Robotic System ( SURS ) for Single Port Laparoscopy,” \textit{Ieee/Asme Trans. Mechatronics}, vol. 20, no. 5, pp. 2133--2145, 2014.

\bibitem{loco}J. Whitman, N. Zevallos, M. Travers, and H. Choset, “Snake Robot Urban Search after the 2017 Mexico City Earthquake,” 2018 \textit{IEEE Int. Symp. Safety, Secur. Rescue Robot. SSRR 2018}, pp. 7--12, 2018, doi: 10.1109/SSRR.2018.8468633.

\bibitem{li_1}Z. Li, J. Feiling, H. Ren, and H. Yu, “A Novel Tele-Operated Flexible Robot Targeted for Minimally Invasive Robotic Surgery,” \textit{Engineering}, vol. 1, no. 1, pp. 073--078, Mar. 2015, doi: 10.15302/J-ENG-2015011.

\bibitem{furet r1}M. Furet, M. Lettl, and P. Wenger, “Kinematic Analysis of Planar Tensegrity 2-X Manipulators,” in \textit{Springer Proceedings in Advanced Robotics}, vol. 8, pp. 153--160, 2019.

\bibitem{wenger r2}P. Wenger and D. Chablat, “Kinetostatic analysis and solution classification of a class of planar tensegrity mechanisms,” \textit{Robotica}, vol. 37, no. 7, pp. 1214--1224, Jul. 2019, doi: 10.1017/S026357471800070X.

\bibitem{li_2} Z. Li, L. Wu, H. Ren, and H. Yu, “Kinematic comparison of surgical tendon-driven manipulators and concentric tube manipulators,” \textit{Mech. Mach. Theory}, vol. 107, no. June 2016, pp. 148-165, 2017, doi: 10.1016/j.mechmachtheory.2016.09.018.

\bibitem{le_1}H. M. Le, T. N. Do, L. Cao, and S. J. Phee, “Towards active variable stiffness manipulators for surgical robots,” \textit{Proc. - IEEE Int. Conf. Robot. Autom.}, pp. 1766--1771, 2017, doi: 10.1109/ICRA.2017.7989209.

\bibitem{shang}J. Shang et al., “A Single-Port Robotic System for Transanal Microsurgery-Design and Validation,” \textit{IEEE Robot. Autom. Lett.}, vol. 2, no. 3, pp. 1510-1517, 2017, doi: 10.1109/LRA.2017.2668461.

\bibitem{haber}G. P. Haber et al., “SPIDER surgical system for urologic procedures with laparoendoscopic single-site surgery: From initial laboratory experience to first clinical application,” \textit{Eur. Urol.}, vol. 61, no. 2, pp. 415--422, 2012, doi: 10.1016/j.eururo.2010.12.033.

\bibitem{abbott}D. J. Abbott, C. Becke, R. I. Rothstein, and W. J. Peine, “Design of an endoluminal NOTES robotic system,” \textit{IEEE Int. Conf. Intell. Robot. Syst.}, pp. 410--416, 2007, doi: 10.1109/IROS.2007.4399536.

\bibitem{roh} S. G. Roh et al., “Development of the SAIT single-port surgical access robot slave arm based on RCM Mechanism,” \textit{Proc. Annu. Int. Conf. IEEE Eng. Med. Biol. Soc. EMBS}, vol. 2015-Novem, pp. 5285--5290, 2015, doi: 10.1109/EMBC.2015.7319584.

\bibitem{hu_1} X. Hu, A. Chen, Y. Luo, C. Zhang, and E. Zhang, “Steerable catheters for minimally invasive surgery: a review and future directions,” \textit{Comput. Assist. Surg.}, vol. 23, no. 1, pp. 21--41, 2018, doi: 10.1080/24699322.2018.1526972.

\bibitem{Akopov2021} A. Akopov, D. Y. Artioukh, and T. F. Molnar, “Surgical
Staplers: The History of Conception and Adoption,” The
Annals of Thoracic Surgery, vol. 112, pp. 1716–1721, nov
2021.

\bibitem{Bolanos1997}H. Bolanos, E. Norwalk;, C. R. Sherts, Southport;, A. T. Pel-
letier, and Wallingford, “ENDOSCOPIC STAPLER,” US
Patent 690269, 1997.

\bibitem{Milliman1999} L. Milliman, Bethel;, F. J. Viola, S. Hook;, J. Orban, Nor-
walk;, R. F. Lehn, and Stratford, “SURGICAL STAPLING
APPARATUS,” US Patent 586531, 1999.

\bibitem{rouillier}
Rouillier, F., “Solving zero-dimensional systems through the rational univariate representation,” Applicable Algebra in Engineering, Communication and Computing, vol.9, no. 5, pp. 433--461, 1999.

\bibitem{behzadipour st1}S. Behzadipour and A. Khajepour, “Stiffness of cable-based parallel manipulators with application to stability analysis,” \textit{J. Mech. Des. Trans. ASME}, vol. 128, no. 1, pp. 303--310, 2006, doi: 10.1115/1.2114890.

\bibitem{chadefaux st2}T. Chadefaux, “The Triggers of War: Disentangling the Spark from the Powder Keg,” \textit{SSRN Electron. J.}, no. April 2014, 2014, doi: 10.2139/ssrn.2409005.

\bibitem{wanda}W. Zhao, A. Pashkevich, A. Klimchik, and D. Chablat, “Elastostatic Modeling of Multi-Link Flexible Manipulator Based on Two-Dimensional Dual-Triangle Tensegrity Mechanism,” \textit{J. Mech. Robot.}, vol. 14, no. 2, pp. 1--31, 2022, doi: 10.1115/1.4051789.

\bibitem{yang_1}J. Yang, E. P. Pitarch, J. Potratz, S. Beck, and K. Abdel-Malek, “Synthesis and analysis of a flexible elephant trunk robot,”\textit{ Adv. Robot.}, vol. 20, no. 6, pp. 631--659, 2006, doi: 10.1163/156855306777361631.


\end{thebibliography}
\end{document}